\documentclass[runningheads]{llncs}
\usepackage[T1]{fontenc}
\usepackage{graphicx}
\usepackage{cite}
\usepackage{amsmath,amssymb,amsfonts}
\usepackage{algorithmic}
\usepackage{textcomp}
\usepackage{xcolor}
\usepackage{subcaption}

\begin{document}
\title{Emergence of Fixational and Saccadic Movements in a Multi-Level Recurrent Attention Model for Vision}
\titlerunning{Emergence of Fixational and Saccadic Movements in MRAM}
%
\author{Pengcheng Pan\inst{1} \and
	Yonekura Shogo\inst{1} \and
	Yasuo Kuniyoshi\inst{1}}
\authorrunning{P. Author et al.}
%
\institute{The University of Tokyo, Tokyo, Japan \\
	\email{\{pan,yonekura,kuniyoshi\}@isi.imi.i.u-tokyo.ac.jp}}
\maketitle              
\begin{abstract}
	Inspired by foveal vision, hard attention models promise interpretability and parameter economy. However, existing models like the Recurrent Model of Visual Attention (RAM) and Deep Recurrent Attention Model (DRAM) failed to model the hierarchy of human vision system, that compromise on the visual exploration dynamics. As a result, they tend to produce attention that are either overly fixational or excessively saccadic, diverging from human eye movement behavior.
	In this paper, we propose a {\bf M}ulti-Level {\bf R}ecurrent {\bf A}ttention {\bf M}odel (MRAM), a novel hard attention framework that explicitly models the neural hierarchy of human visual processing. By decoupling the function of glimpse location generation and task execution in two recurrent layers, MRAM emergent a balanced behavior between fixation and saccadic movement. Our results show that MRAM not only achieves more human-like attention dynamics, but also consistently outperforms CNN, RAM and DRAM baselines on standard image classification benchmarks.
	
\keywords{ Image classification \and Active Vision \and Recurrent Neural Networks \and Hard Attention \and Biologically Inspired Computation}
\end{abstract}

\section{Introduction}
\label{sec:intro}
Human vision actively interacts with the environment by eye movement, characterized by recurrent fixations on specific spatial locations and saccadic movements. Rather than passively receiving uniform visual input, human eyes deploy focused glimpses of high-resolution processing in the centered fovea vision, complemented by peripheral.
This gazed information integrates over time, allowing humans to understand and explore relevant scene regions in a sequential, yet highly flexible manner.
Seminal work by Yarbus demonstrates that human eye scanpaths, formed by sequences of fixations and saccades, reflect not only the viewer’s focus but also mental activities such as the nature of the task \cite{Yarbus}. Although factors such as the ongoing tasks and the expert knowledge of humans have an impact on the scan paths, human eye movement shares similarities in the fixational and saccadic patterns \cite{Eye1, Eye2}.

Studies in neuroscience reveal the cortical pathway of human visual information processing. Human visual processing includes two distinct pathways from the retina, one being the geniculostriate pathway, that information going through the lateral geniculate nucleus (LGN) to the primary visual cortex, and another, the eye movement pathway, directly going to the superior colliculus (SC) along the optic nerve, projecting to the pons paramedian pontine reticular formation that sends control signals to oculomotor nuclei controlling the eye movement \cite{Kandel}. The processed information in the primary visual cortex, posterior parietal cortex, and frontal eye fields also projects back to the SC for eye movement control. It was shown by lesions of the visual cortex causing a loss of peripheral vision while the center vision remains unaffected. This suggests that the center vision relies mainly on the visual information directly received by the retina to the SC for eye movement control, while the geniculostriate pathway processes the center vision and peripheral vision and provides feedback for indirect eye movement control.

Recent advances in machine learning and computational methods of visual modeling focus on the whole image processing. The Convolutional Neural Network (CNN) model has shown high accuracy on visual tasks and similarity in the visual information processing in the visual cortex \cite{CNN}. However, such computation usually has a high computation cost and a black-box model that is insufficient in interpreting its prediction. In addressing these problems, attention models in machine learning seek to replicate the nature of foveal vision and attention mechanisms. While soft attention employs a standard CNN to distribute the attention map over the entire input, hard attention models employ a learnable policy to select small sub-regions or “glimpses” and process them with higher resolution. The hard attention strategy can reduce computational demand, improve interpretability, and learn a dynamic human-like gaze behavior. Even though hard attention is harder to train than soft attention because the cut glimpses are non-differentiable. Seminal work by Mnih et al. proposed the Recurrent Model of Visual Attention (RAM), showing one can learn hard attention through the Reinforcement Learning (RL) method \cite{RAM}. They train a recurrent network to sequentially choose glimpse locations and to perform tasks such as classification from partial observations with the REINFORCE algorithm \cite{REINFORCE}. The subsequent variations, such as the Deep Recurrent Attention Model (DRAM), are proposed to solve dedicated visual tasks such as multi-object recognition.

Nevertheless, to the best of our knowledge, most existing hard attention frameworks, such as RAM and DRAM, are designed to replicate the human attention roughly or through the conventional geniculostriate pathway that most machine learning models in vision applied.
In addition, these models focus on the task performance, such as classification accuracy and glimpse over the object, while their glimpse movement patterns, including fixations or saccades, are not well studied. As a consequence, existing methods may struggle to balance the learning for task performance and for a good policy in governing the glimpse location choice.

In this work, we introduce the Multi-level Recurrent Attention Model (MRAM), a novel hard attention approach motivated by the hierarchical nature of the human eye movement pathway (Section~\ref{sec:method}). 
Through experiments on benchmark datasets on MNIST, FashionMNIST and FER2013, we demonstrate that MRAM improves object classification performance while human-like fixations and saccades emerge (Section~\ref{sec:experiments}).

\section{Related Work}
\label{sec:related}

\subsection{Recurrent Models of Visual Attention}
Visual attention has long stood as a cornerstone topic in computer vision research. Studies from neuroscience indicate that signals guided by prior knowledge interact with those influenced by raw input, creating a competitive process to select relevant elements in a scene \cite{attention1, attention2, biased1, biased2}. Early efforts in this area introduced saliency-based approaches, usually determined by local low-level feature contrast \cite{saliency}. Incorporating feedback neural networks, recurrent networks, and reinforcement learning into attention-based networks has led to noteworthy gains across a variety of visual tasks \cite{attention3, attention4, attention5, RLattention}.

Recent methods evolved to allocate processing resources more efficiently. For instance, hard attention strategies limit computation to specific parts of an image, reducing computational demands \cite{fovea, Xu}. The RAM then broadened this idea, using sequential glimpses to examine visual content while preserving interpretability \cite{RAM}.  Following RAM, the DRAM was proposed, which used multiple glimpses in deeper recurrent structures, showing improvements on multi-digit recognition tasks such as the Street View House Numbers dataset \cite{DRAM}. More recent variants of RAM extended the hard attention to dynamically adjusting the stop time, modeling the saccadic behavior by restraining the model from visited locations, and a biological approach using predictive coding \cite{DyRAM, saccader, PCRAM}.

\subsection{Hierarchical and Multi-Scale Extensions}

Hierarchical strategies have gained attention due to the layered nature of tasks and the need to accommodate targets at varying levels of detail. One early demonstration of this principle introduced a tiered feature extraction pipeline, successfully capturing both low-level cues and higher-level semantics for object recognition and segmentation \cite{hierarchy1}. Meanwhile, modeling how people look around in the context of goals and sub-goals also highlights the importance of structured, task-driven control of eye movements \cite{hierarchy2}.

Within deep learning, sequential models have embraced multi-scale designs that split complex input into progressively more abstract representations. For instance, a layered recurrent network has been shown to improve skeletal action recognition by separately capturing local movements before integrating them globally \cite{HRNN1}. A related design relies on an attention mechanism stacked at different levels—focusing first on words, then on sentences—to boost performance in document classification \cite{HAN}. Extensions of this theme push further into multi-resolution contexts, dealing effectively with intricate temporal or spatial patterns \cite{HRNN2, HRNN3}.

Beyond supervised approaches, reinforcement learning has also taken advantage of hierarchical structures. Subdividing a large goal into manageable chunks can accelerate training and produce more robust behaviors \cite{HRL}. In general, these hierarchical and multi-scale frameworks have shown notable improvements in tackling visual challenges, reflecting the real-world complexity of tasks and their multi-level requirements. Therefore, a hierarchical RAM model aligns more closely with the way humans move their eyes while also leveraging the tiered nature of visual and task demands, ultimately enhancing performance in our approach.

\section{Method}
\label{sec:method} 
\begin{figure}[htbp]
	\centerline{\includegraphics[scale=0.5]{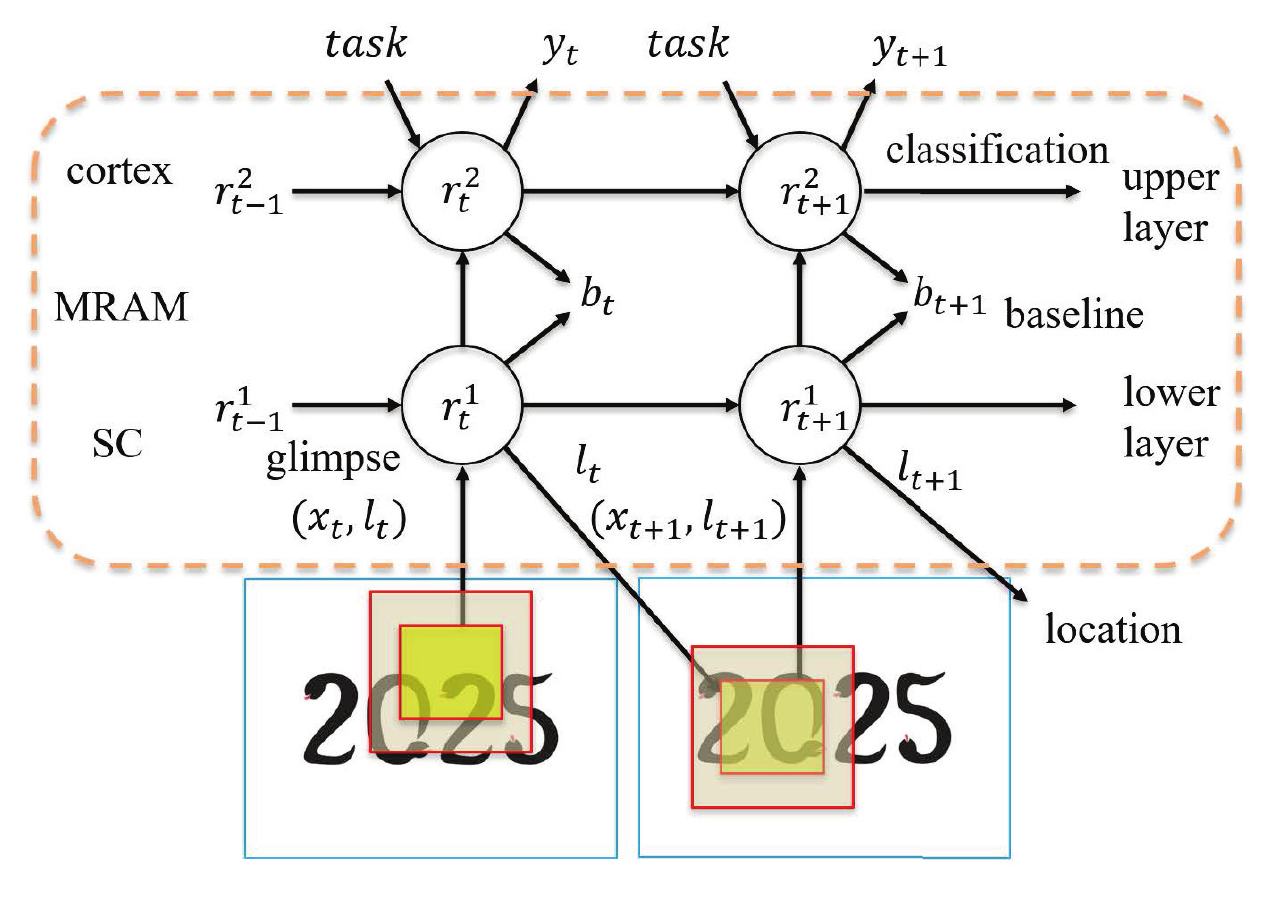}}
	\caption{The multi-layer recurrent attention model.}
	\label{fig1}
\end{figure}

Our approach employs the RAM, extending it to a hierarchical formulation in which multiple recurrent layers cooperate to attend to glimpsed regions and to produce sequential classification decisions. DRAM was designed for the multiple object recognition with a similar multi-layers structure. The primary differences in our method are (1) we modeled the human eye movement pathway so that the top (higher) layer is responsible for classification while the bottom (lower) layer learns to predict the next glimpse location, and (2) we modeled the feedback projection from the primary visual cortex to SC in the baseline method used in the policy gradient training so that the baseline is computed from a hybrid hidden state of both layers. The overall process is illustrated in Fig.~\ref{fig1}. 

\subsection{RAM} 
The RAM utilizes RL for a Partially Observed Markov Decision Process (POMDP) problem setting. At each time step $t$, the model works as an agent that actively interacts with the scene by placing the partially observed glimpse while making predictions on the image class. The RAM can be formulated in the following 4 modules:
\begin{enumerate} 
	\item \textbf{Glimpse module.} A glimpse network, parameterized by $\mathcal{G}_t $, that concatenates the glimpse location $\ell_t$ and the patched glimpse image $x_t$, a small region around $\ell_t$, and a down-sampled larger region in the same location with the scale factor scales $s$, using linear layer $\theta_g$, from the input image: 
	\begin{equation}
		\mathcal{G}_t = f_{g}\bigl(\left[x_t\;\bigl\|\;x^s_t\right],\ell_t; \theta_g\bigr)\label{eq1}
	\end{equation}
	\item \textbf{Core module.} A core network, parameterized by $\mathcal{H}_t$, that utilizes a Recurrent Neural Network (RNN) model for processing the sequential glimpse $\mathcal{G}_t $ across time:
	\begin{equation}
		\mathcal{H}_t = f_{h}\bigl(H_{t-1}, \mathcal{G}_t; \theta_h\bigr)\label{eq2}
	\end{equation}
	\item \textbf{Location module.} The location network, parameterized by $\mathcal{L}_t $, utilizes a Multilayer Perceptron (MLP) to predict the next glimpse location $\ell_{t+1}$ from the hidden state $H_{t}$:
	\begin{equation}
		\mathcal{L}_t = f_{l}\bigl(H_t; \theta_l\bigr)\label{eq3}
	\end{equation}
	
	\item \textbf{Action module.} The action network uses another MLP to execute the image classification task of predicting the image class $a_t$ from the hidden state $H_{t}$:
	\begin{equation}
		\mathcal{\alpha}_t = f_{a}\bigl(H_t; \theta_a\bigr)\label{eq4}
	\end{equation}
\end{enumerate}

\subsection{MRAM} 
We introduce two recurrent layers, with the Long-Short Term Memory (LSTM) model \cite{LSTM} serving as the RNN at each layer, denoted $h^1_t$ and $h^2_t$. At each time step $t$: 

\begin{enumerate} 
	
	\item \textbf{Lower recurrent layer ($H^1_t$).} Like the human eye movement pathway, the lower layer governs the glimpse location updates while receiving the glimpsed image as input directly. The lower layer acts as a pure RAM model. It receives the glimpsed image $\mathcal{G}_t$ and the previous hidden state $H^1_{t-1}$. The local glimpse features are processed, and the recurrent outputs are passed to the location module and a higher recurrent layer.
	\begin{equation}
		\mathcal{H}^{1}_t = f_{h_{1}}\bigl(H^{1}_{t-1}, \mathcal{G}_t; \theta^{1}_h\bigr)\label{eq5}
	\end{equation}
	\item \textbf{Higher recurrent layer ($H^2_t$).} The top layer acts like the primary visual cortex that integrates for an abstract object representation while receiving indirect visual information from the lower layer. It takes the recurrent outputs from $H^1_t$, as well as its previous hidden state $H^2_{t-1}$. A high-level visual representation is learned, and the recurrent outputs are passed to the action module for image classification.
	\begin{equation}
		\mathcal{H}^{2}_t = f_{h_{2}}\bigl(H^{2}_{t-1}, \mathcal{H}^{1}_t; \theta^{2}_h\bigr)\label{eq6}
	\end{equation}
	\item \textbf{Hierarchical factorisation of POMDP.} In MRAM, we model visual search as a partially–observable MDP with latent state
	$s_t$, observation $o_t$, gaze action $g_t$, and terminal classification $y$.
	In a \emph{flat} RAM, a single RNN must learn
	the full joint policy $\pi_\theta(g_{1:T},y\!\mid\!o_{1:T})$,
	entangling short–horizon saccade control with long–horizon recognition. 
	In DRAM, the lower RNN learn long–horizon recognition, while the higher RNN learn the short–horizon saccade control. 
	The mismatch in time-scale leads to compromised performance on both gaze action and classification.
	We instead factorise the policy into a two‑layer hierarchy
	(SC–layer ${\pi_L}$, VC–layer ${\pi_H}$):
	
	\begin{equation}
		\pi_\theta(g_{1:T},y \mid o_{1:T})= \prod_{t=1}^{T} \pi_L\!\bigl(g_t \mid h^{1}_{t-1}\bigr)\;\times\;\pi_H\!\bigl(y \mid h^{2}_{T}\bigr)\label{eq7}
	\end{equation}
	
	where \(\pi_L\) evolves on a \emph{fast} time–scale (saccadic glimpse control,
	superior colliculus) and \(\pi_H\) on a \emph{slow} time–scale
	(object recognition, visual cortex).
	The factorisation is lossless yet separates exploration (\(\pi_L\)) from exploitation (\(\pi_H\)),
	reducing credit‑assignment span like typical hierarchical reinforcement learning method did.

\end{enumerate} 

\subsection{Hybrid Baseline Estimation} 
To reduce variance in policy gradient training, we applied a baseline function $b_t = \mathop{\mathbb{E}}_\pi[R_t]$ that approximates the expected return \cite{baseline}. Unlike standard baselines in RAM or DRAM, which often use a single hidden state, our hybrid baseline fuses the hidden states from both layers, resembling the feedback projection from the primary visual cortex to SC: 
\begin{equation}
	b_t = f_{\mathrm{b}}\bigl(\left[h^1_t\;\bigl\|\;h^2_t\right]; \theta_b\bigr)\label{eq8}
\end{equation}
where “\(\|\)” denotes concatenation and $f_{\mathrm{b}}$ is a small multi-layer perceptron producing the scalar baseline. This design leverages both low-level and higher-level cues, providing a more stable estimate for the reward expectation.

\subsection{Training} 
We follow a policy gradient approach for learning the location policy akin to the same REINFORCE algorithm applied in RAM \cite{RAM, REINFORCE}. The location policy for the lower layer is a stochastic component, which samples the glimpse location $\ell_{t+1}$ from a distribution parameterized by $h^1_t$. The action module of the higher layer is learned in a supervised approach with the classification outputs $y_t$ and the ground-truth label. The overall hybrid loss combines: 
\begin{enumerate} 
	\item A classification loss, computed with the cross-entropy comparing $y_T$ and the ground-truth label. 
	\item A policy gradient loss by REINFORCE, which maximizes the expected reward. The reward is given 1 if the classification is correct, 0 otherwise. We subtract $b_t$ to reduce variance: 
	\begin{equation}
		\mathcal{L}_{\mathrm{REINFORCE}} = -\sum_{t=1}^T (R - b_t) \log \pi(\ell_t \mid h^1_t; \theta), \label{eq6}
	\end{equation}
	where $R$ is the reward and $\pi(\ell_t\mid h^1_t; \theta)$ is the policy distribution based on hidden state of the lower recurrent layer $H^1_t$. 
\end{enumerate} 
Hence, the total objective is: 
\begin{equation}
	\mathcal{L} \;=\; \mathcal{L}_{\text{classification}} +\mathcal{L}_{\text{baseline}} + \alpha \,\mathcal{L}_{\mathrm{REINFORCE}},\label{eq7}
\end{equation}
where $\alpha$ balances these 3 terms. 

\section{Experiments}
\label{sec:experiments}

We conduct experiments on three different standard image classification task to illustrate the efficacy of MRAM. For fair comparison, we tested the LeNet-5 CNN, RAM and DRAM with a simple CNN backbone on the same device as baselines. All hyperparameters including glimpse sizes, number of scales, and number of glimpses are held constant for comparison across methods for fairness, unless otherwise indicated. The RNN is a standard 256 LSTM units for all RAM-based model and both layers. In training, the parameter $\alpha$ is set to be 0.01, the random seed fixed to 1, batch size fixed to 128, total epoch fixed to 300 with early stopping when model is on a plateau for 50 epochs, the Adam optimizer is used with initial learning rate $3 \times 10^{-4}$, which will be deduced when validation accuracy has stopped improving. All experiment does not include data augmentation, and only normalization is used for data pre-processing, We performed all experiments on a single NVIDIA RTX 3070 graphic board.

\subsection{MNIST}
\label{sec:mnist}

In this task, we evaluate the model in same standard dataset used in RAM, MNIST datasets. The MNIST dataset is handwritten images ($28 \times 28$) with digital numbers ranging from 0 to 9 \cite{MNIST}. We evaluate the contribution of each module to MRAM with ablation on the CNN and baseline. We record the performance of different models and the effectiveness of individual modules such as total trainable parameters and average training time of a single epoch. We compare our MRAM model to RAM and DRAM under identical training conditions, with a small LeNet-5 as CNN baseline, reporting test set accuracy in Tables~\ref{tab1}.

\begin{table}[htbp]
	\caption{Classification result on MNIST}
	\begin{center}
		\resizebox{\textwidth}{!}{%
			\begin{tabular}{|l|c|c|c|}
				\hline
				Model & \multicolumn{1}{c|}{Params (M)} & \multicolumn{1}{c|}{Infer Time (ms/im)} & Accuracy\\
				\hline
				\hline
				CNN(LeNet-5) & 0.061 & 2.60 ± 0.15 & 99.04\%\\
				\hline
				RAM, 7 glimpses & 0.637 & 2.16 ± 0.09 & 99.05\%\\
				\hline
				RAM, 10 glimpses & 0.637 & 1.75 ± 0.03 & 98.69\%\\
				\hline
				DRAM, 7 glimpses & 3.542 & 2.15 ± 0.02 & 98.63\%\\
				\hline
				DRAM, 10 glimpse & 1.985 & 2.00 ± 0.06 & 99.06\%\\
				\hline
				DRAM without CNN, 7 glimpses  & 1.163 & 2.03 ± 0.01 & 98.16\%\\
				\hline
				DRAM without CNN, 10 glimpses  & 1.163 & 1.95 ± 0.03 & 98.44\%\\
				\hline
				DRAM with hybrid baseline, 10 glimpses  & 1.163 & 2.01 ± 0.03 & 99\%\\
				\hline
				MRAM, 7 glimpses & 1.163 & 1.76 ± 0.04 & 99.2\%\\
				\hline
				MRAM single baseline, 10 glimpses & 1.163 & 1.80 ± 0.05 & \text{99.01}\%\\
				\hline
				MRAM, 10 glimpses & 1.163 & 1.88 ± 0.14 & \textbf{99.26}\%\\
				\hline
			\end{tabular}%
		}
		\label{tab1}
	\end{center}
\end{table}

While RAM is reported the best performance at the 7 glimpse steps, we tested the glimpses to 7 and 10 for comparing short and long glimpse. The glimpsed image size is set to be fixed at $8 \times 8$ same as reported in RAM\cite{RAM}. The number of scale is set to 1.
MRAM achieves a state-of-the-art hard attention accuracy of 99.2\% at 10 glimpse steps, outperforming all baseline models. We also evaluated the contribution of the context, a full image global information extracted by a CNN and passed to the initial state of the higher layer in DRAM as presented in DRAM\cite{DRAM}. The DRAM relies on the context information, and removing the context module leads to a 0.47\% accuracy loss. To justify the contribution of the hybrid baseline method in MRAM, we tested the MRAM train with a single baseline method, the same as the RAM and DRAM models, and the DRAM to train with the hybrid baseline method in the 10 glimpses experiment setting. The MRAM loses 0.25\% accuracy, while DRAM loses 0.06\% accuracy. This suggests the effectiveness of the hybrid baseline method inspired by the feedback projection from the visual cortex to the SC fit only on the MRAM model. Longer glimpse step leads to longer training cost. The cost per glimpse for of RAM/MRAM in training is about 11s/epoch, which is smaller than DRAM (13s/epoch) and CNN, while the inference time doesn't vary much(Tab.\ref{tab1}).

RAM have been tested to show a decreased performance in a longer glimpse step setting. However, while the image recognition could finish in small steps for both human and hard attention models, Yarbus' experiments have shown that humans tend to preserve and recurrently execute the task over a long period. We observed 0.36\% performance drop of RAM in a longer exploration setting, while DRAM and MRAM gained a 0.43\% and 0.06\% performance boost, benefited from multi-layer structure.

\subsection{FashionMNIST}
\label{sec:fashionmnist}
The FashionMNIST dataset is MNIST-like grayscale images ($28 \times 28$) with pictures of common clothes categories such as dress and sneaker \cite{FashionMNIST}. To further explore the performance on image classification and the learned glimpse policy, we record the parameters, inference time per image, accuracy of different models and the predicted locations on the dataset. The baseline, all parameters are the same with MNIST experiment, except we increased the longer glimpse step from 10 to 12 steps for larger gap between short and long glimpses.

\begin{table}[htbp]
	\caption{Classification result on FashionMNIST}
	\begin{center}
		\resizebox{\textwidth}{!}{%
			\begin{tabular}{|l|c|c|c|}
				\hline
				Model & \multicolumn{1}{c|}{Params (M)} & \multicolumn{1}{c|}{Infer Time (ms/im)} & Accuracy\\
				\hline
				\hline
				CNN(LeNet-5) & 0.061 & 1.22 ± 0.21 &  89.02\%\\
				\hline
				RAM, 7 glimpses & 0.637 & 2.14 ± 0.10 & 88.68\%\\
				\hline
				RAM, 12 glimpses & 0.637 & 1.79 ± 0.03 & 90.19\%\\
				\hline
				DRAM, 7 glimpses & 1.985 &  2.16 ± 0.02  & 89.28\%\\
				\hline
				DRAM, 12 glimpses & 1.985 & 1.81 ± 0.02 & 89.56\%\\
				\hline
				DRAM without CNN, 7 glimpses  & 1.163 & 2.13 ± 0.05 & 88.64\%\\
				\hline
				DRAM without CNN, 12 glimpses & 1.163 & 1.75 ± 0.05 & 87.21\%\\
				\hline
				MRAM, 7 glimpses & 1.163 & 2.08 ± 0.03 & \text{91.68}\%\\
				\hline
				MRAM, 12 glimpses & 1.163 & 1.80 ± 0.02 & \textbf{91.79}\%\\
				\hline
			\end{tabular}%
		}
		\label{tab2}
	\end{center}
\end{table}

Table~\ref{tab2} shows that MRAM with 12 glimpses attains 91.79\% accuracy, beating all baselines. The results are similar to MNIST, extending to 12 glimpses yields a minor improvement for both DRAM and MRAM, but also for the RAM, suggesting the longer glimpse contribute to the harder task. DRAM has the largest inference time cost, while RAM and MRAM has a larger cost than CNN.

\subsection{Self-Emergence of Fixations and Saccades in Category Classification}
\label{sec:fixations_saccades}

\begin{figure}[htbp]
	\centerline{\includegraphics[scale=0.4]{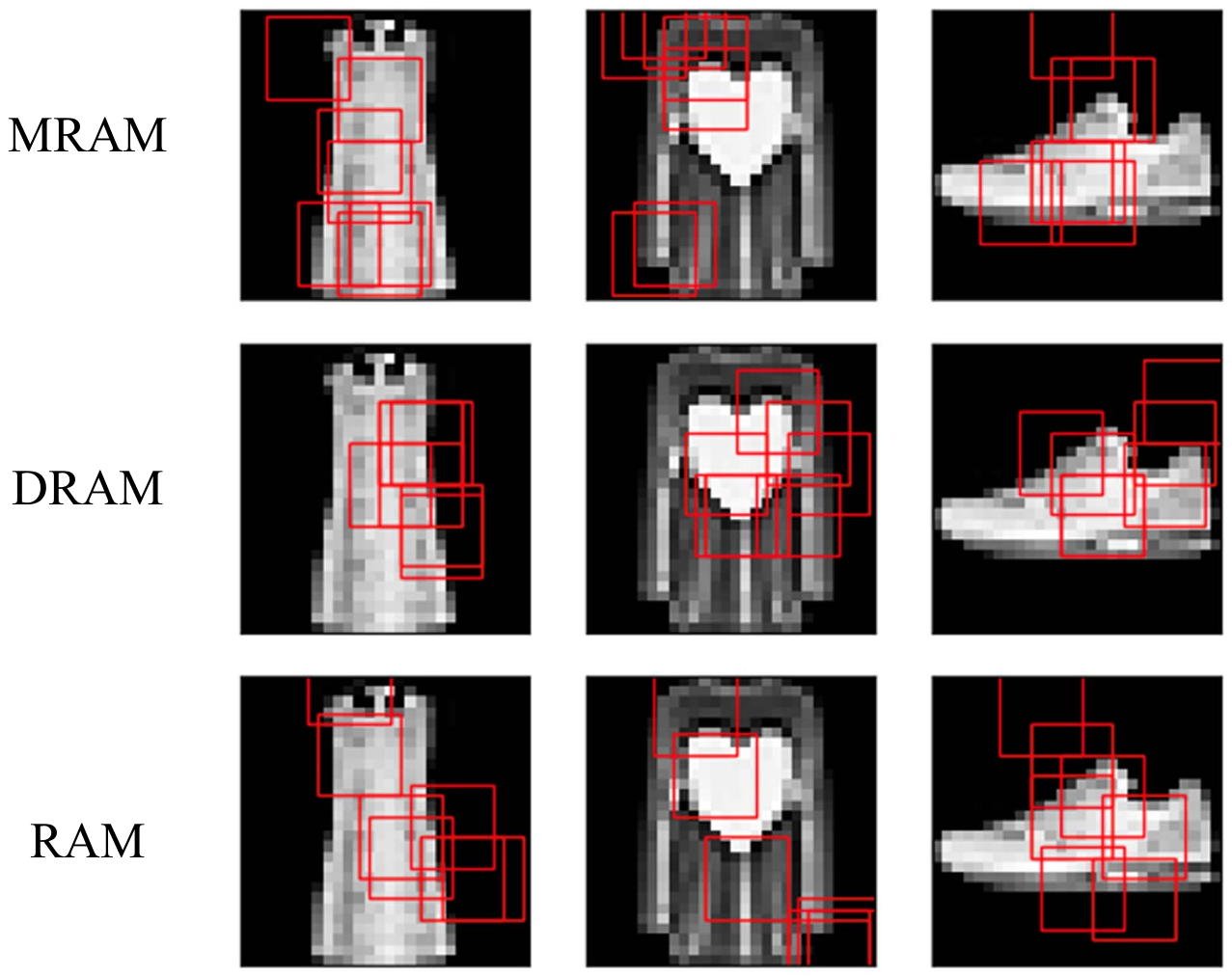}}
	\caption{\textbf{Comparison of learned policies on FashionMNIST.} Red rectangles show the actual glimpse image served as input to the attention model. The top row corresponds to MRAM, the middle row corresponds to DRAM, and the bottom row corresponds to RAM.}
	\label{fig2}
\end{figure}

\begin{figure}[htbp]
	\centering
	\begin{subfigure}{0.47\textwidth}
		\centering
		\includegraphics[width=\linewidth]{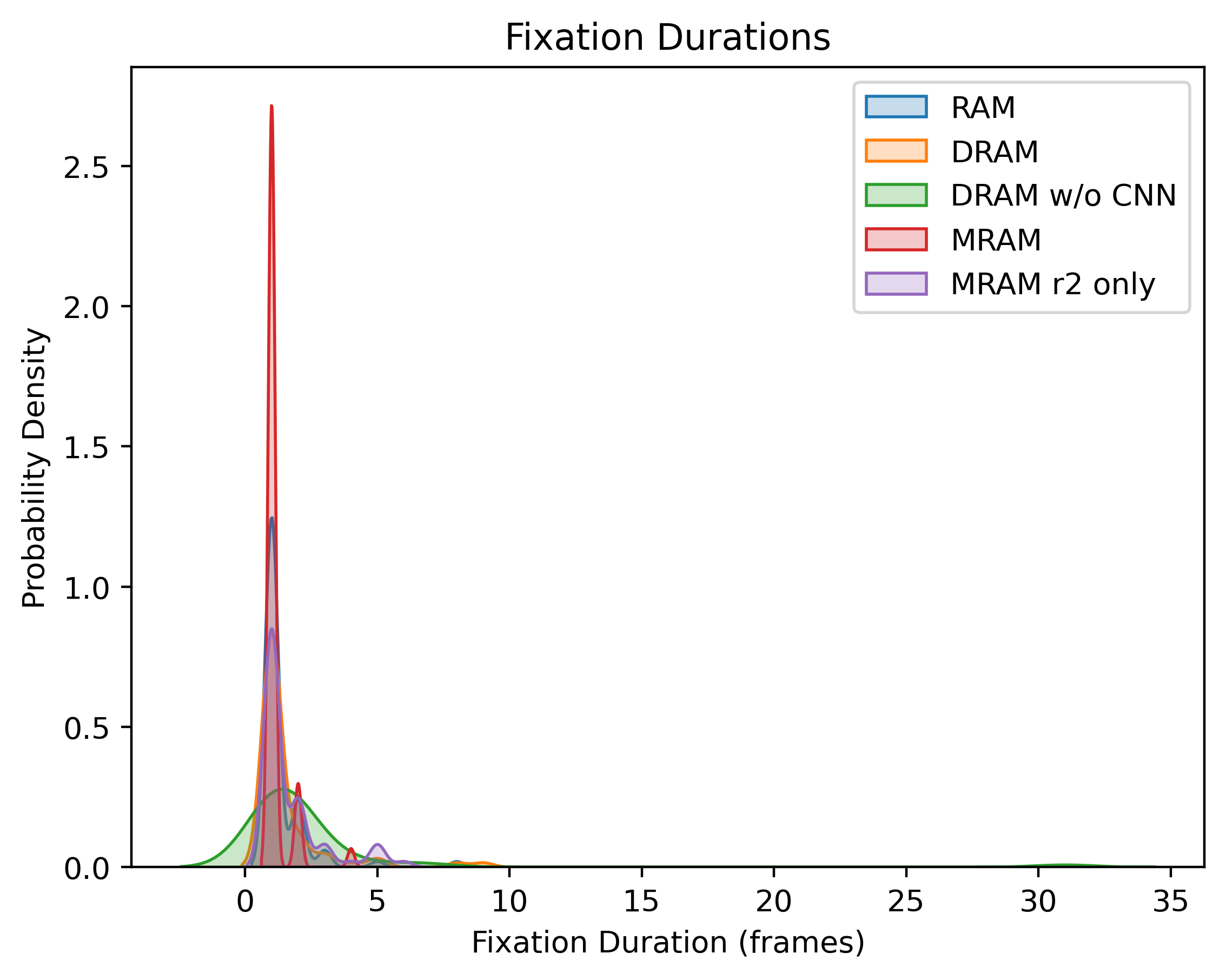}
		\caption{} 
		\label{fig31}
	\end{subfigure}
	\hfill
	\begin{subfigure}{0.47\textwidth}
		\centering
		\includegraphics[width=\linewidth]{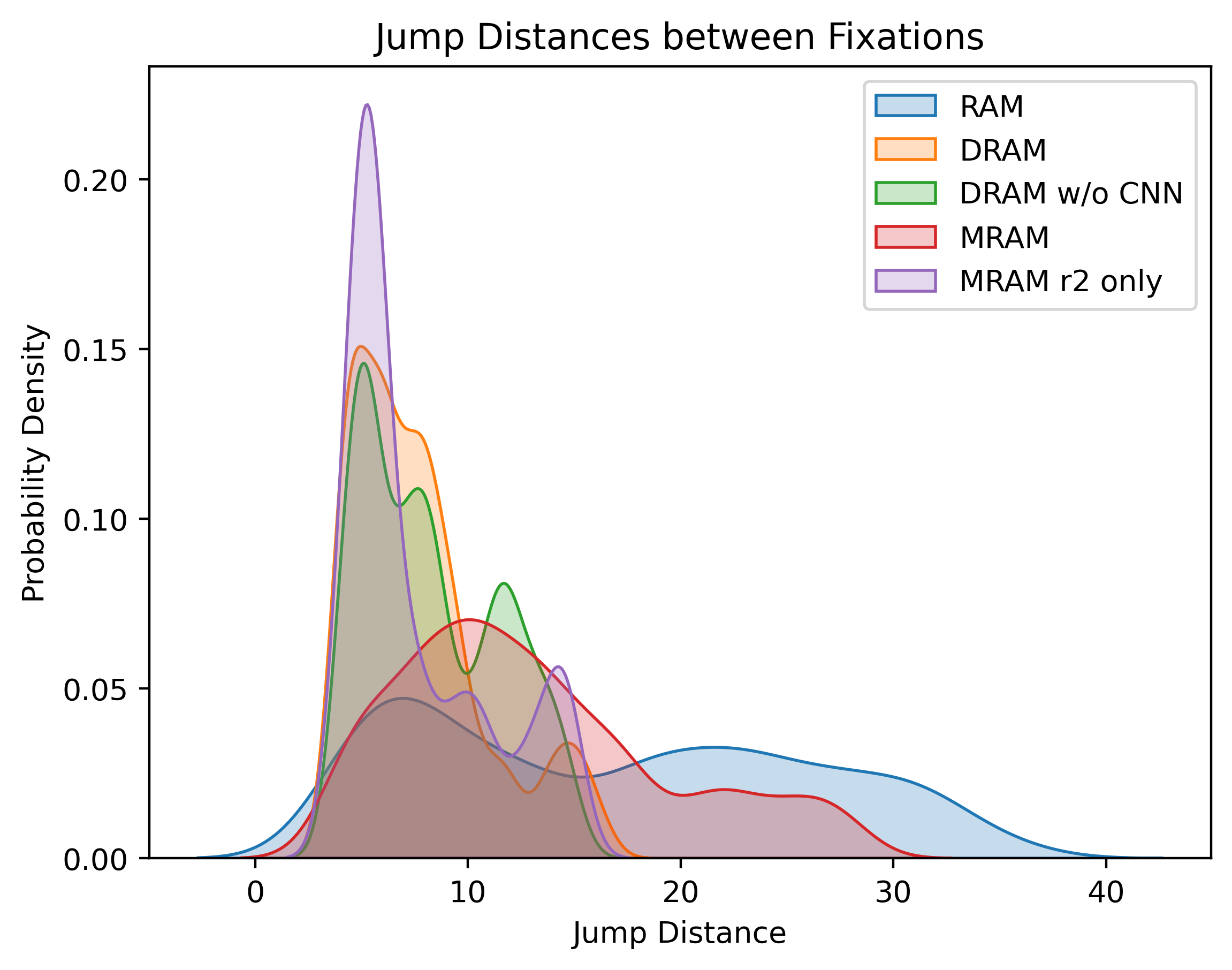}
		\caption{} 
		\label{fig32}
	\end{subfigure}
	\caption{\textbf{Analysis of fixation duration distribution and saccade distance distribution using Kernel Distance Estimate (KDE).} (a) shows the fixation duration distribution of all models in FashionMNIST. (b) shows the saccade distance distribution of all models in FashionMNIST.}
	\label{fig3}
\end{figure}

To answer the question of how MRAM outperformed RAM and whether a human-like gaze policy is learned, we examine the sequentially predicted glimpse locations during the classification task in FashionMNIST.
Figure~\ref{fig2} illustrates examples of the predicted fixation locations for different item categories in FashionMNIST. Notably, MRAM often employs repeated fixations around distinctive features of an object and occasionally executes longer saccades to explore new regions.
In contrast, the RAM model and DRAM model learn a distinct glimpse policy: one learns to continuously place glimpses in new locations like saccades, while the other learns to focus on a relatively small region like fixation. The concentrated glimpse policy suggested DRAM failed to learn the proper gaze policy because of the overwhelmingly powerful global context provided by CNN.
RAM typically reach peak performance at a specific number of glimpses, after which additional steps may provide diminishing returns or even degrade performance. This observation has motivated variants such as Dynamic RAM, which adaptively adjust the number of glimpses \cite{DyRAM}. It can be explained by our visualization result that RAM is always predicting gaze like saccade, and a manual stop is necessary before the saccade moved away from target. In contrast, our MRAM framework can sustain or improve its classification accuracy as the number of glimpses increases, leveraging its hierarchical structure to incorporate both fixations and broader saccadic shifts. In general, MRAM’s hierarchical design, inspired by the human brain, enables it to maintain robustness by refining local information across glimpses while also making larger saccades when needed.

To quantitatively explain the visualization of the learned policy of each model, we showed the probability density of fixation durations and jump distance with the kernel distance estimate on the predicted glimpse locations. As the glimpse windows are $8 \times 8$, it's natural to define a fixation by the distance of two glimpses lower than 8. We defined the fixation with a more strict threshold distance of 6. The probability density of fixation durations and jump distances is shown in Figure~\ref{fig3}. MRAM learned a glimpse policy with short fixation durations but longer jump distances compared to DRAM.
This emerged pattern echoes the findings of Yarbus \cite{Yarbus} and the fixations duration and saccade distance data of human vision in \cite{Saccadepattern1}, in which humans combine long-time fixations and sporadic long-range gaze shifts. Intriguingly, we reproduced the DRAM-like behavior on MRAM, when the locator uses only output from the higher recurrent layer during test(MRAM r2 only). This ablation support that human-like eye movement comes from our hierarchical structure decoupling 2 RNN layers for 2 distinct functions.
Categories such as dresses and shirts, which span a significant vertical portion of the image, benefit especially from multiple fixations capturing various sections of the item’s silhouette. Our method thus learns to focus on salient or discriminative regions repeatedly while also retaining the flexibility to jump to new areas when needed, ultimately leading to improved recognition performance.

\subsection{FER2013}
\label{sec:FER2013}
\begin{table}[htbp]
	\caption{Classification result on FER}
	\begin{center}
		\begin{tabular}{|l|c|c|c|}
			\hline
			Model & \multicolumn{1}{c|}{Params (M)} & \multicolumn{1}{c|}{Infer Time (ms/im)} & Accuracy\\
			\hline
			\hline
			CNN(LeNet-5) & 0.158 & 5.0 ± 2.15 & 48.4\%\\
			\hline
			RAM, 1 scale & 0.637 & 4.99 ± 1.90 & 42.81\%\\
			\hline
			RAM, 2 scale & 0.637 & 5.50 ± 1.71 & 46.27\%\\
			\hline
			DRAM, 1 scale & 1.985 & 4.73 ± 1.32 & 48.38\%\\
			\hline
			DRAM, 2 scale & 1.985 & 5.76 ± 1.80 & 42.63\%\\
			\hline
			DRAM without CNN, 1 scale  & 1.163 & 4.27 ± 1.42 & 48.69\%\\
			\hline
			DRAM without CNN, 2 scale  & 1.163 & 5.42 ± 1.66 & 42.84\%\\
			\hline
			MRAM, 1 scale & 1.163  & 4.45 ± 1.37 & \textbf{48.73}\%\\
			\hline
			MRAM, 2 scale & 1.163 & 5.47 ± 1.68 & \textbf{48.04\%}\\
			\hline
		\end{tabular}
		\label{tab3}
	\end{center}
\end{table}

To evaluate our model on real-world image, we tested our models and baseline models on the Facial Expression Recognition (FER) 2013 dateset with ($48 \times 48$) images of human face\cite{FER}. As we tested that longer glimpse is beneficial for all models in FashionMNIST, we fix the glimpse steps to 12, and the glimpsed image size is set to be fixed at $8 \times 8$. In the 2 scale setting, the peripheral vision of $16 \times 16$ image patch will be down-sampled to $8 \times 8$, and concatenated as the image input to the RAM-based models.

\begin{figure}[htbp]
	\centerline{\includegraphics[scale=0.35]{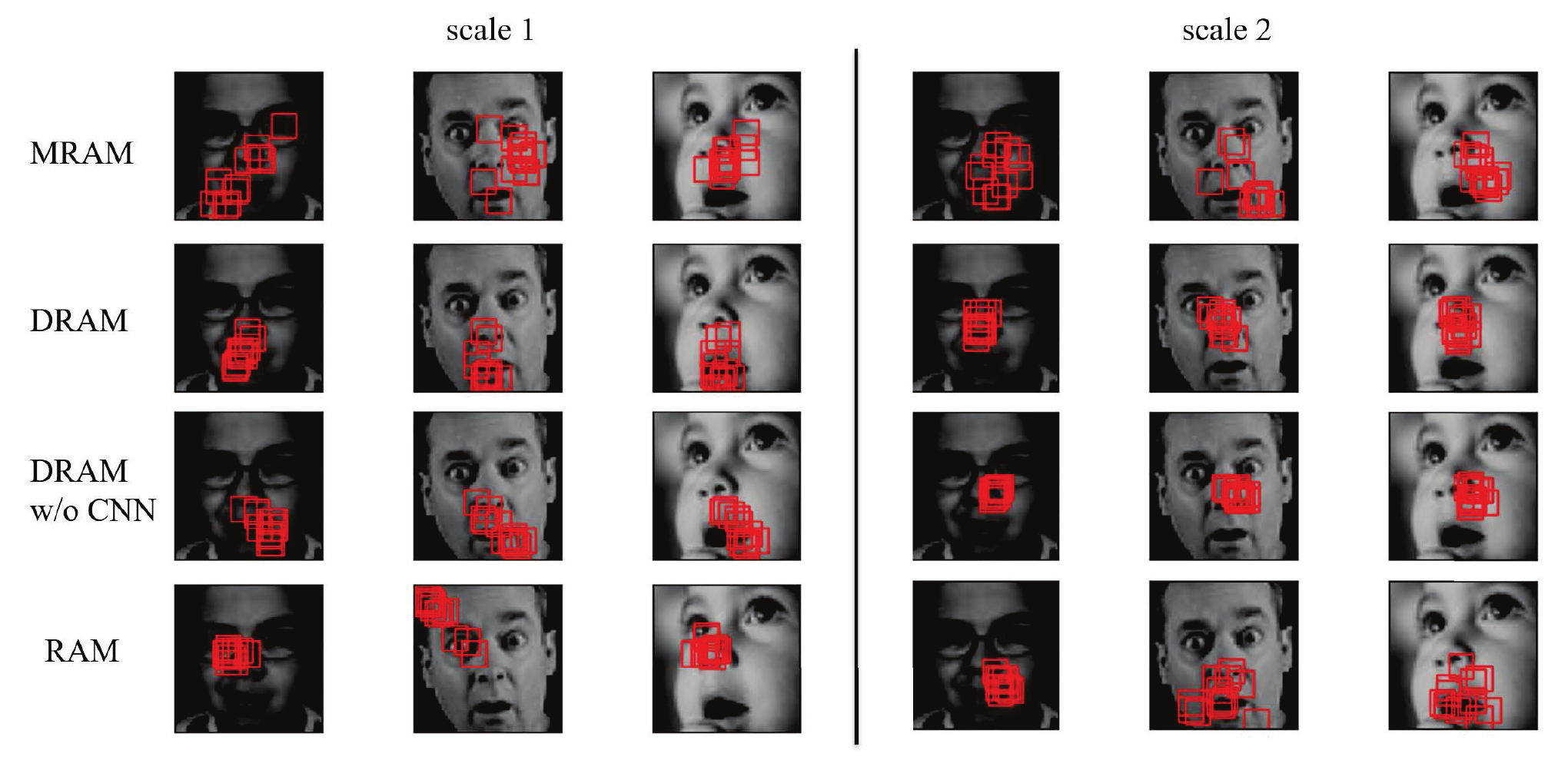}}
	\caption{Examples of the learned policy on the FER2013 classification task.}
	\label{fig4}
\end{figure}

Table~\ref{tab3} reports classification accuracy for MRAM and baselines. MRAM achieves the best performance (48.73\% acc.) in it the 1 scale configuration. When increasing the scale, all RAM-based model except RAM loses accuracy. To answer this reason, figure~\ref{fig4} shows visualization samples of the learned policies for RAM-based models. Notably, only MRAM learned to put glimpse to eye, mouth, and nose of human face, while all other models failed to learn the gaze policy properly. When scale is increased, MRAM failed to learn to put gaze that cover all key locations of face, resulting in 0.69\% decreased accuracy. Same to the FashionMNIST experiment, DRAM is affected by the global context provided by CNN, that removing the CNN will lead to higher accuracy. With broader vision in 2 scale, RAM learned different gaze policy to fixation on mouth, which is more informative than other facial regions, and this policy leads to high classification accuracy. There is no much difference in the inference cost across models.

\section{Discussion}
\label{sec:discussion}

This study presents the MRAM, which is inspired by the human eye movement pathway. Our model extends the traditional RAM by incorporating a two-layer recurrent structure that separately handles glimpse policy and image classification.
The results from our experiments demonstrate that MRAM not only performs competitively against state-of-the-art hard attention models like DRAM but also exhibits several advantages in terms of its ability to model human-like attention patterns and its robustness to increasing the number of glimpses.

One of the key strengths of MRAM lies in its ability to produce emergent fixation and saccade patterns resembling human vision. MRAM takes inspiration from the human neural pathway of visual processing.
As shown in Figures~\ref{fig2}, \ref{fig3}, and \ref{fig4}, MRAM learns to alternate between short-range fixations for detail refinement and longer-range saccades for exploring new areas.
These results align with Yarbus' observations of human visual behavior \cite{Yarbus}, and statistics data of human eye movement \cite{Saccadepattern1}, where fixations and long saccades emerge naturally during visual exploration.
This behavior suggests that a hierarchical attention model can naturally approximate the dynamics of eye movement with repetitive fixation and exploratory saccade.
Such patterns not only improve interpretability but also hint at the potential for MRAM to generalize well to other visual tasks that require more complex attention strategies.
From a broader perspective, the experiment results reinforce the idea that the designed hierarchical attention mechanisms are not only computationally effective but also biologically plausible.
The key limitations of MRAM lies in the unbalanced peripheral vision, that leads to the performance drop in the 2 scale setting, which mimic human visual system closer. The ablation study of CNN in DRAM suggested human brain have additional architecture to balance between recognition and gaze policy, that can be applied to more biological plausible hard attention model.

\section{Conclusion}
\label{sec:conclusion}

In this work, we presented the MRAM, a novel multi-layer hard attention framework inspired by the multi-scale nature of human visual processing. 
We introduce a two-layer hierarchy to RAM, where the lower layer focuses on selecting informative glimpses and the upper layer is dedicated to producing classification decisions and a hybrid baseline method that provide feedback from the upper layer to the lower layer. 
Our experiments demonstrated advantages of our model over lightweight CNN, RAM and DRAM baselines on the MNIST, FashionMNIST and FER2013 datasets while producing more human-like attention patterns, characterized by emergent fixations and saccades.
Future work will explore a more human-like visual attention model integrating CNN and peripheral vision in MRAM and extend the scalability to more diverse and complex datasets.
We believe MRAM provides a solid foundation for developing next-generation human-like attention models that are not only powerful and efficient, but also interpretable and closer to human vision.

\begin{credits}
\subsubsection{\ackname}This study was funded by JST SPRING GX project (Grant Number JPMJSP2108). This preprint has not undergone peer review (when applicable) or any post-submission improvements or corrections. The Version of Record of this contribution is published in Neural Information Processing. ICONIP 2025. Lecture Notes in Computer Science, vol 16310, and is available online at https://doi.org/10.1007/978-981-95-4378-6\_21

\subsubsection{\discintname}
The authors have no competing interests to declare that are
relevant to the content of this article.
\end{credits}
%
%
%
\bibliographystyle{splncs04}
\bibliography{mybibliography}

@String{Computer = "{IEEE} Computer" }

@String{Springer = "Springer-Verlag" }

@misc{FER,
	author = {Dumitru and Ian Goodfellow and Will Cukierski and Yoshua Bengio},
	title = {Challenges in Representation Learning: Facial Expression Recognition Challenge},
	year = {2013},
	note = {Kaggle}
}

@book{Yarbus,
   author = {Alfred L. Yarbus},
   journal = {Eye Movements and Vision},
   publisher = {Springer US},
   title = {Eye Movements and Vision},
   year = {1967}
}

@book{Kandel,
  author = {Kandel, Eric R. and Schwarz, James H. and Jessel, Thomas M.},
  edition = {4.},
  publisher = {McGraw-Hill},
  title = {Principles of {N}eural {S}cience},
  year = 2000
}

@article{CNN,
  author = {LeCun, Yann and Bengio, Yoshua and Hinton, Geoffrey},
  journal = {nature},
  number = 7553,
  pages = 436,
  publisher = {Nature Publishing Group},
  title = {Deep learning},
  volume = 521,
  year = 2015
}

@article{saliency,
  author = {Itti, Laurent and Koch, Christof and Niebur, Ernst},
  journal = {IEEE Transactions on pattern analysis and machine intelligence},
  number = 11,
  pages = {1254--1259},
  publisher = {Citeseer},
  title = {{A model of saliency-based visual attention for rapid scene analysis}},
  volume = 20,
  year = 1998
}

@inproceedings{Xu,
 author       = {Kelvin Xu and
                  Jimmy Ba and
                  Ryan Kiros and
                  Kyunghyun Cho and
                  Aaron C. Courville and
                  Ruslan Salakhutdinov and
                  Richard S. Zemel and
                  Yoshua Bengio},
  editor       = {Francis R. Bach and
                  David M. Blei},
  title        = {Show, Attend and Tell: Neural Image Caption Generation with Visual
                  Attention},
  booktitle    = {Proceedings of the 32nd International Conference on Machine Learning,
                  {ICML} 2015, Lille, France, 6-11 July 2015},
  series       = {{JMLR} Workshop and Conference Proceedings},
  volume       = {37},
  pages        = {2048--2057},
  year         = {2015},
}

@inproceedings{RAM,
  author = {Mnih, Volodymyr and Heess, Nicolas and Graves, Alex and Kavukcuoglu, Koray},
  booktitle = {NIPS},
  editor = {Ghahramani, Zoubin and Welling, Max and Cortes, Corinna and Lawrence, Neil D. and Weinberger, Kilian Q.},
  pages = {2204-2212},
  title = {Recurrent Models of Visual Attention.},
  year = 2014
}

@inproceedings{DRAM,
  author = {Ba, Jimmy and Mnih, Volodymyr and Kavukcuoglu, Koray},
  description = {Multiple Object Recognition with Visual Attention},
  booktitle    = {3rd International Conference on Learning Representations, {ICLR} 2015,
                  San Diego, CA, USA, May 7-9, 2015, Conference Track Proceedings},
  title = {Multiple Object Recognition with Visual Attention},
  year = 2014
}

@inproceedings{saccader,
  author = {Elsayed, Gamaleldin F. and Kornblith, Simon and Le, Quoc V.},
  booktitle = {NeurIPS},
  editor = {Wallach, Hanna M. and Larochelle, Hugo and Beygelzimer, Alina and d'Alché Buc, Florence and Fox, Emily B. and Garnett, Roman},
  ee = {http://papers.nips.cc/paper/8359-saccader-improving-accuracy-of-hard-attention-models-for-vision},
  pages = {700-712},
  title = {Saccader: Improving Accuracy of Hard Attention Models for Vision.},
  year = 2019
}

@inproceedings{DyRAM,
  author = {Li, Zhichao and Yang, Yi and Liu, Xiao and Zhou, Feng and Wen, Shilei and Xu, Wei},
  booktitle = {ICCV Workshops},
  ee = {https://doi.ieeecomputersociety.org/10.1109/ICCVW.2017.145},
  isbn = {978-1-5386-1034-3},
  pages = {1199-1209},
  title = {Dynamic Computational Time for Visual Attention.},
  year = 2017
}

@article{PCRAM,
  author = {Sharafeldin, Abdelrahman and Imam, Nabil and Choi, Hannah},
  ee = {https://doi.org/10.1016/j.patter.2024.100983},
  journal = {Patterns},
  number = 6,
  pages = 100983,
  title = {Active sensing with predictive coding and uncertainty minimization.},
  volume = 5,
  year = 2024
}

@article{Eye1,
  author = {Rayner, Keith},
  issn = {0033-2909},
  journal = {Psychological Bulletin},
  number = 3,
  pages = {p372 - 422},
  title = {Eye movements in reading and information processing: 20 years of research.},
  volume = 124,
  year = 1998
}

@article{Eye2,
title = {Eye movements in natural behavior},
journal = {Trends in Cognitive Sciences},
volume = {9},
number = {4},
pages = {188-194},
year = {2005},
issn = {1364-6613},
author = {Mary Hayhoe and Dana Ballard}
}

@InProceedings{Saccadepattern1,
author="Gajewski, Daniel A.
and Pearson, Aaron M.
and Mack, Michael L.
and Bartlett, Francis N.
and Henderson, John M.",
editor="Paletta, Lucas
and Tsotsos, John K.
and Rome, Erich
and Humphreys, Glyn",
title="Human Gaze Control in Real World Search",
booktitle="Attention and Performance in Computational Vision",
year="2005",
publisher="Springer Berlin Heidelberg",
address="Berlin, Heidelberg",
pages="83--99",
isbn="978-3-540-30572-9"
}

@article{attention1,
  author = {Desimone, R. and Duncan, J.},
  description = {Neural mechanisms of selective visual attention},
  journal = {Annual Review of Neuroscience},
  pages = {193-222--},
  refid = {1212},
  title = {Neural mechanisms of selective visual attention},
  username = {jabreftest},
  volume = 18,
  year = 1995
}

@article{attention2,
  author = {Butko, Javier R. Movellan Nicholas J.},
  description = {Human Detection},
  journal = {Computer Vision and Pattern Recognition, 2009. CVPR 2009. Proceedings
	of the 2009 IEEE Computer Society Conference on},
  title = {Optimal Scanning for Faster Object Detection},
  year = 2009
}

@inproceedings{attention3,
  author = {Cao, Chunshui and Liu, Xianming and Yang, Yi and Yu, Yinan and Wang, Jiang and Wang, Zilei and Huang, Yongzhen and Wang, Liang and Huang, Chang and Xu, Wei and Ramanan, Deva and Huang, Thomas S.},
  booktitle = {ICCV},
  ee = {https://doi.ieeecomputersociety.org/10.1109/ICCV.2015.338},
  isbn = {978-1-4673-8391-2},
  pages = {2956-2964},
  publisher = {IEEE Computer Society},
  title = {Look and Think Twice: Capturing Top-Down Visual Attention with Feedback Convolutional Neural Networks.},
  year = 2015
}

@inproceedings{attention4,
  author = {Fu, Jianlong and Zheng, Heliang and Mei, Tao},
  booktitle = {CVPR},
  ee = {https://doi.ieeecomputersociety.org/10.1109/CVPR.2017.476},
  isbn = {978-1-5386-0457-1},
  pages = {4476-4484},
  publisher = {IEEE Computer Society},
  title = {Look Closer to See Better: Recurrent Attention Convolutional Neural Network for Fine-Grained Image Recognition.},
  year = 2017
}

@inproceedings{attention5,
  author = {Qiao, Tingting and Dong, Jianfeng and Xu, Duanqing},
  booktitle = {AAAI},
  editor = {McIlraith, Sheila A. and Weinberger, Kilian Q.},
  ee = {https://doi.org/10.1609/aaai.v32i1.12272},
  pages = {7300-7307},
  publisher = {AAAI Press},
  title = {Exploring Human-Like Attention Supervision in Visual Question Answering.},
  year = 2018
}

@inproceedings{RLattention,
  author = {Caicedo, Juan C. and Lazebnik, Svetlana},
  booktitle = {ICCV},
  ee = {https://doi.ieeecomputersociety.org/10.1109/ICCV.2015.286},
  isbn = {978-1-4673-8391-2},
  pages = {2488-2496},
  publisher = {IEEE Computer Society},
  title = {Active Object Localization with Deep Reinforcement Learning.},
  year = 2015
}

@inproceedings{fovea,
  author = {Larochelle, Hugo and Hinton, Geoffrey E.},
  booktitle = {NIPS},
  editor = {Lafferty, John D. and Williams, Christopher K. I. and Shawe-Taylor, John and Zemel, Richard S. and Culotta, Aron},
  ee = {http://papers.nips.cc/paper/4089-learning-to-combine-foveal-glimpses-with-a-third-order-boltzmann-machine},
  pages = {1243-1251},
  publisher = {Curran Associates, Inc.},
  title = {Learning to combine foveal glimpses with a third-order Boltzmann machine.},
  year = 2010
}

@inproceedings{hierarchy1,
  author = {Girshick, R. and Donahue, J. and Darrell, T. and Malik, J.},
  booktitle = {2014 IEEE Conference on Computer Vision and Pattern Recognition (CVPR)},
  description = {Rich Feature Hierarchies for Accurate Object Detection and Semantic Segmentation},
  issn = {1063-6919},
  month = {June},
  pages = {580-587},
  title = {Rich Feature Hierarchies for Accurate Object Detection and Semantic Segmentation},
  volume = 00,
  year = 2014
}

@article{hierarchy2,
   author = {Mary Hayhoe and Dana Ballard},
   issn = {09609822},
   issue = {13},
   journal = {Current Biology},
   month = {7},
   pages = {R622-R628},
   pmid = {25004371},
   publisher = {Cell Press},
   title = {Modeling task control of eye movements},
   volume = {24},
   year = {2014}
}

@inproceedings{HRNN1,
  author = {Du, Yong and Wang, Wei and Wang, Liang},
  booktitle = {CVPR},
  ee = {https://doi.ieeecomputersociety.org/10.1109/CVPR.2015.7298714},
  isbn = {978-1-4673-6964-0},
  pages = {1110-1118},
  publisher = {IEEE Computer Society},
  title = {Hierarchical recurrent neural network for skeleton based action recognition.},
  year = 2015
}

@inproceedings{HAN,
  address = {San Diego, California},
  author = {Yang, Zichao and Yang, Diyi and Dyer, Chris and He, Xiaodong and Smola, Alex and Hovy, Eduard},
  booktitle = {Proceedings of the 2016 Conference of the NAACL-HLT},
  description = {Hierarchical Attention Networks for Document Classification - ACL Anthology},
  editor = {Knight, Kevin and Nenkova, Ani and Rambow, Owen},
  pages = {1480--1489},
  publisher = {Association for Computational Linguistics},
  title = {Hierarchical Attention Networks for Document Classification},
  year = 2016
}

@Inbook{HRL,
author={Hengst, Bernhard},
title={Hierarchical Reinforcement Learning},
bookTitle={Encyclopedia of Machine Learning},
year={2010},
publisher={Springer US},
address={Boston, MA},
pages={495--502},
isbn={978-0-387-30164-8},
}

@inproceedings{HRNN2,
  title        = {5th International Conference on Learning Representations, {ICLR} 2017,
                  Toulon, France, April 24-26},
  author = {Chung, Junyoung and Ahn, Sungjin and Bengio, Yoshua},
  description = {[1609.01704] Hierarchical Multiscale Recurrent Neural Networks},
  title = {Hierarchical Multiscale Recurrent Neural Networks},
  year = {2017}
}

@inproceedings{HRNN3,
  author = {Degtyarenko, Illya and Deriuga, Ivan and Grygoriev, Andrii and Polotskyi, Serhii and Melnyk, Volodymyr and Zakharchuk, Dmytro and Radyvonenko, Olga},
  booktitle = {ICASSP},
  ee = {https://doi.org/10.1109/ICASSP39728.2021.9413412},
  isbn = {978-1-7281-7605-5},
  pages = {2865-2869},
  publisher = {IEEE},
  title = {Hierarchical Recurrent Neural Network for Handwritten Strokes Classification.},
  year = 2021
}

@article{biased1,
title = {Top-down and bottom-up mechanisms in biasing competition in the human brain},
journal = {Vision Research},
volume = {49},
number = {10},
pages = {1154-1165},
year = {2009},
issn = {0042-6989},
author = {Diane M. Beck and Sabine Kastner}
}

@article{biased2,
author = {Desimone, Robert},
year = {1998},
month = {09},
pages = {1245-55},
title = {Visual attention mediated by biased competition in extrastriate visual cortex},
volume = {353},
journal = {Philosophical transactions of the Royal Society of London. Series B, Biological sciences},
}

@article{LSTM,
  author = {Hochreiter, Sepp and Schmidhuber, J{\"u}rgen},
  description = {Weiterführende Literatur zu Long Short-Term Memory},
  journal = {Neural computation},
  number = 8,
  pages = {1735--1780},
  publisher = {MIT Press},
  title = {Long short-term memory},
  volume = 9,
  year = 1997
}

@online{FashionMNIST,
  author       = {Han Xiao and Kashif Rasul and Roland Vollgraf},
  title        = {Fashion-MNIST: a Novel Image Dataset for Benchmarking Machine Learning Algorithms},
  date         = {2017-08-28},
  year         = {2017},
  eprintclass  = {cs.LG},
  eprinttype   = {arXiv},
  eprint       = {cs.LG/1708.07747},
}

@article{MNIST,
  title={The mnist database of handwritten digit images for machine learning research},
  author={Deng, Li},
  journal={IEEE Signal Processing Magazine},
  volume={29},
  number={6},
  pages={141--142},
  year={2012},
  publisher={IEEE}
}

@article{REINFORCE,
  author = {Williams, R. J.},
  journal = {Machine Learning},
  pages = {229--256},
  title = {Simple statistical gradient-following algorithms for connectionist reinforcement learning},
  volume = 8,
  year = 1992
}

@inproceedings{baseline,
  acmid = {3009806},
  address = {Cambridge, MA, USA},
  author = {Sutton, Richard S. and McAllester, David and Singh, Satinder and Mansour, Yishay},
  booktitle = {Proceedings of the 12th International Conference on Neural Information Processing Systems},
  description = {Policy gradient methods for reinforcement learning with function approximation},
  location = {Denver, CO},
  numpages = {7},
  pages = {1057--1063},
  publisher = {MIT Press},
  series = {NIPS'99},
  title = {Policy Gradient Methods for Reinforcement Learning with Function Approximation},
  year = 1999
}
\end{document}